\pdfoutput=1
\documentclass[11pt]{article}

\usepackage[]{acl}

\usepackage{times}
\usepackage{latexsym}
\usepackage{enumitem}
\usepackage{hyperref}
\usepackage{tikz}
\usetikzlibrary{positioning}
\usepackage{amssymb} 
\usepackage[T1]{fontenc}
\usepackage[utf8]{inputenc}
\usepackage{multirow}
\usepackage{microtype}
\usepackage{xurl}
\usepackage{graphicx}
\usepackage{xcolor}
\PassOptionsToPackage{dvipsnames,svgnames,table}{xcolor}
\usepackage{array}
\usepackage{booktabs}
\definecolor{green}{RGB}{153,255,153}
\definecolor{aqua}{RGB}{162,232,211}
\definecolor{hred}{RGB}{255,153,153}
\definecolor{darkblue}{RGB}{0,0,153}
\definecolor{teal}{RGB}{0,153,153}
\usepackage{soul}
\DeclareRobustCommand{\hlgreen}[1]{\sethlcolor{green}\hl{#1}}
\DeclareRobustCommand{\hlred}[1]{\sethlcolor{hred}\hl{#1}}

\usepackage{caption}
\usepackage{subcaption}
\newcommand{\compress}{\vspace{-2ex}}
\usepackage{enumitem}
\title{Concept-Based Explanations to Test for False Causal Relationships Learned by Abusive Language Classifiers }
\author{ Isar Nejadgholi, Svetlana Kiritchenko, Kathleen C. Fraser, and Esma Balk{\i}r\\
  National Research Council Canada \\
  Ottawa, Canada \\
 \footnotesize \texttt{\{Isar.Nejadgholi,Svetlana.Kiritchenko,Kathleen.Fraser,Esma.Balkir\}@nrc-cnrc.gc.ca}\\
 }

\begin{document}
\maketitle
\begin{abstract}
Classifiers tend to learn a false causal relationship between an over-represented concept and a label, which can result in over-reliance on the concept and compromised classification accuracy. It is imperative to have methods in place that can compare different models and identify over-reliances on specific concepts. We consider three well-known abusive language classifiers trained on large English datasets and focus on the concept of \textit{negative emotions}, which is an important signal but should not be learned as a sufficient feature for the label of abuse. Motivated by the definition of \textit{global sufficiency}, we first examine the unwanted dependencies learned by the classifiers by assessing their accuracy on a challenge set across all decision thresholds. Further, recognizing that a challenge set might not always be available, we introduce concept-based explanation metrics to assess the influence of the concept on the labels. These explanations allow us to compare classifiers regarding the degree of false global sufficiency they have learned between a concept and a label. 

{\textit{\textbf{Content Warning:} This paper presents examples that may be offensive or upsetting.}}
\end{abstract}

\section{Introduction}
\label{sec:intro}
In various natural language classification tasks, particularly in abusive language detection, certain concepts are known to be strong signals for the label of interest. These concepts are often over-represented in the respective class of the training set, making them susceptible to being learned as potential causes for the label. Consequently, the classifier over-relies on these concepts and ignores the broader context, leading to reduced generalizability \cite{yin2021towards}. Hence, to ensure models are robust and reliable, it is crucial to develop methods that can detect these over-reliances in various natural language classification tasks.

\begin{figure}
     \centering
     \includegraphics[trim={0cm 0 0 0cm}, clip,width=0.45\textwidth]{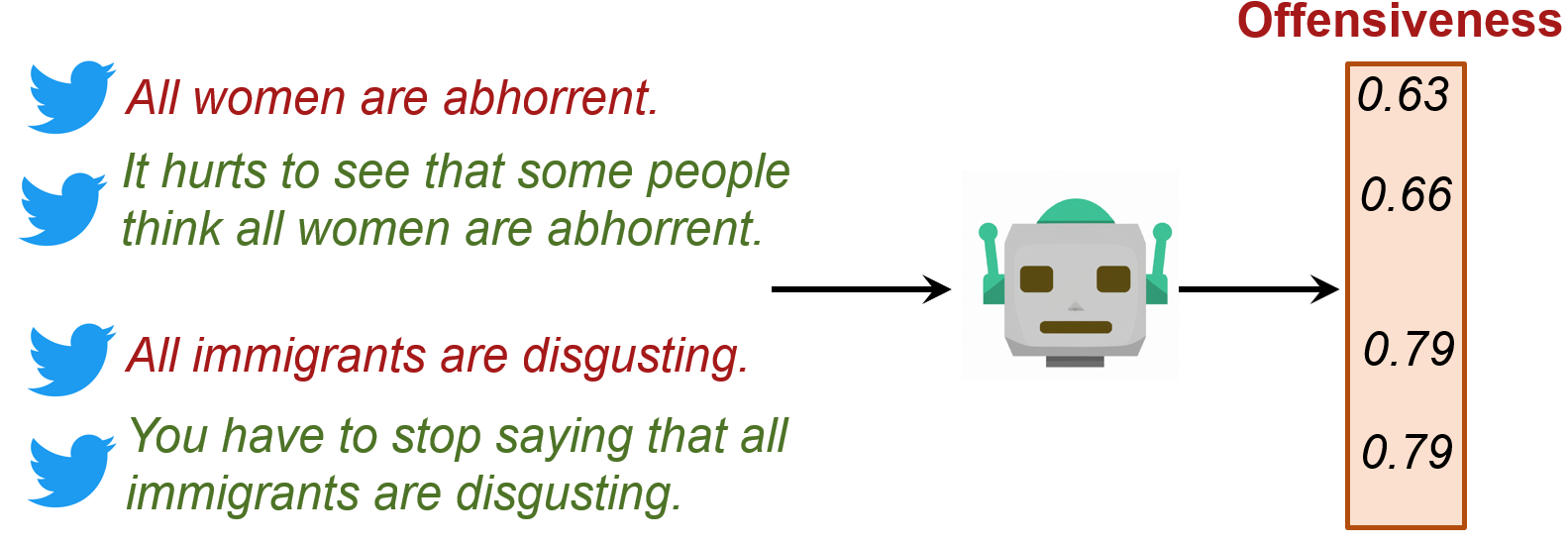}
    %\vspace{-10pt}
     \caption{Probability of offensiveness generated by the TweetEval  Classifier \cite{barbieri-etal-2020-tweeteval}. The classifier has learned a false global sufficiency between negative emotions and the label of offense. It over-relies on this concept and ignores the broader context. }
        \label{fig:senario}
         %\vspace{-10pt}
\end{figure}

In the context of abusive language detection, we consider the concept of \textit{negative emotions}. The presence of an expression %evoking or
associated with \textit{negative emotion} is an important signal for detecting abusive language and has been used in feature-based systems before \cite{chiril2022emotionally,fortuna2018survey}. Crucially, in some examples, negative emotion words might be the cause for the abusive label, i.e., the sentence might not be abusive if the negative emotion word is replaced with other words (e.g., \textit{I know these people. They are \textbf{disgusting}}). However, at the global level, the relationship between negative emotion and abusive language is a strong correlation, not causation, as it is neither globally necessary nor globally sufficient for the label of abuse.\footnote{Phenomenon P is globally sufficient for the Phenomenon Q, if whenever P happens, Q happens too. P is globally necessary for Q if whenever Q happens, P happens, too \cite{zaeem2021cause}.} Negative emotions are not globally necessary for the label of abuse because there are abusive sentences that do not contain any negative emotion words (e.g., offensive jokes, stereotyping and microaggressions). Also, words evoking negative emotions are not globally sufficient for a sentence to be abusive when interpreted in a broader context (e.g., \textit{We should admit that in our society, they are \textbf{oppressed}.}). But, an end-to-end model might learn that negative emotion in a sentence is globally sufficient for that sentence to be abusive. Such a classifier will struggle in classifying non-abusive sentences that contain negative emotion words leading to a lack of generalizability. An example of such a case is shown in Figure~\ref{fig:senario}. Specifically, classifiers' over-reliance on the negative emotion signal can inadvertently discriminate against marginalized groups since their communications (e.g., discussing their experiences of discrimination and marginalization) can contain negative emotion words and, therefore can be wrongly considered abusive.

We explore a scenario where a user, aware of the importance of negative emotions for their use case, wants to evaluate and compare a set of trained models. Their goal is to identify and eliminate those models that are more prone to generating inaccurate results due to an overemphasis on negative emotions as primary indicators. For that, we use concept-based explanations to test if a model has learned a false global causal relationship between a user-identified concept and a label where the true relationship is a correlation. Note that global causal relationships explain the model's output across an entire dataset, as opposed to local causal explanations, which concern the dependency of an individual prediction on a specific input feature.

%Explainable AI (XAI) offers the potential to make machine learning models safer and more transparent by providing insights into the inner workings of black box models \cite{belle2021principles}. Explanations are most helpful when they are faithful to the reasoning behind decisions made by an algorithm. To that end, the machine learning community has employed the concepts of necessity and sufficiency from the causality literature to understand the cause of decisions made by machine learning algorithms \cite{watson2021local,balkir-etal-2022-necessity,joshi2022all}.  

Concept-based explanations are a class of explainability methods that provide global explanations at the level of human-understandable concepts \cite{hitzler2022human}. While local explanations help the users understand the model's reasoning for an individual decision with respect to the input features, global explanations are critical in comparing the processes learned by models and selecting the one that best suits the needs of a use case \cite{balkir-etal-2022-challenges,burkart2021survey}. Global explanations might be obtained at the level of input features through aggregating local explanations \cite{lundberg2020local}. Alternatively, global-by-design methods (e.g., probing classifiers \cite{conneau2018you}) can be used to gain insights at higher levels of abstractions, such as linguistic properties or human-defined concepts.\footnote{Here, we use the term “feature” to refer to the latent representations of a semantic concept learned by a classifier.}

Similar to most feature importance explainability methods (e.g, \citet{ribeiro2016should,lundberg2017A}), concept-based explanations are originally designed to measure the \textit{importance} of a concept. The intuitive meaning of \textit{importance } usually refers to correlation, and it can be interpreted differently based on two notions of causality: necessity and sufficiency \cite{galhotra2021explaining}. Local explainability methods usually focus on features that are of high local necessity or high local sufficiency for the label \cite{watson2021local,balkir-etal-2022-necessity,joshi2022all}, thus considered important by human users. However, at the global level, all features must be interpreted in a larger context for accurate decision-making. %In end-to-end classifiers, where we do not control how the model learns the relationships between features and labels, the model can learn a false causation relationship (high global sufficiency) between a label and concepts over-represented in the corresponding class of the training set.
We aim to determine if concept-based explanations can be utilized to evaluate whether a trained binary classifier for abusive language detection has learned a false global sufficiency relationship between the label and the concept of negative emotion.
%Here, we ask whether concept-based explanations can be used to assess if a trained binary abusive language classifier has learned a false global sufficiency relationship between the concept of negative emotion and a label. (
Our code and data are available at {\footnotesize \url{https://github.com/IsarNejad/Global-Sufficiency/tree/main}}. Our main contributions are:
%\compress
\begin{itemize}[leftmargin=*]

\item We formalize the issue of over-reliance on a concept as falsely learned global sufficiency. For the task of an abusive language classifier, we consider concepts related to \textit{negative emotion} as being important but not globally sufficient for the label of abuse. We discuss how learning these concepts as globally sufficient results in compromised classification accuracies. 
%For three abusive language classifiers trained with large English datasets and two variations of concepts related to \textit{negative emotion}, we show that one of the classifiers over-relies on emotion-related concepts significantly more than the other two classifiers. 

\item Based on our formalization of false global sufficiency, as a baseline method, we measure the over-reliance of models on a human-defined concept using an unseen challenge set that contains the concept in both classes. Recognizing that various classifiers may have a distinct range of optimal decision thresholds, we assess the over-reliance on a concept across all possible decision thresholds and show that one of the classifiers over-relies on emotion-related concepts significantly more than the other two classifiers. 

\item Taking the challenge set approach as a baseline for comparison, we propose novel concept-based explanation metrics, demonstrating that similar conclusions about the degree of false global sufficiency can be drawn using these metrics. Building on previous work, we modify the TCAV procedure to measure not only the feature's importance but also the extent of its impact on the label. We conclude that a concept-based method is preferable as it eliminates the need for manual data curation.  
\end{itemize}

% \begin{figure*}
%     \centering
%     \small
%     \begin{tikzpicture}
%     %nodes
%     \node (easyprob){\includegraphics[width=.25\textwidth]{images/easy_prob.png}}
%     \node (easyauc){\includegraphics[width=.25\textwidth]{images/easy_auc.png}}

% \end{tikzpicture}
% \caption{}
%  \label{fig:illusteration}
% \end{figure*}

% \section{Emotion Lexicons }
% \label{sec:lexicon}
% \begin{itemize}
% \item intensity emotion lexicon
% \item intensity higher than >0.5
% \end{itemize}

\section{Concept-Based Explanations}
\label{sec:TCAV}
%Concept-based explanations promise to explain neural network models at the abstraction level of human-defined concepts \cite{hitzler2022human}. While most explainability methods provide importance scores for input tokens \citep{sundararajan2017axiomatic, smilkov2017smoothgrad, selvaraju2017grad, shrikumar2017learning}, 

Concept-based explanations evaluate the model’s decision-making mechanism at the level of a human-defined concept expected to be important for the task \cite{koh2020concept}. Specifically, we use the Testing Concept Activation Vectors (TCAV) method to measure the influence of a human-defined concept on the model's predictions \cite{kim2018interpretability}. The idea of TCAV is based on the observation that human-understandable concepts can be encoded as meaningful and insightful information in the linear vector space of trained neural networks \cite{mikolov2013efficient}. A Concept Activation Vector (CAV), which represents the concept in the embedding space, is a vector normal to a hyperplane that separates concept and non-concept examples. Such a hyperplane is obtained by training a linear binary classifier to separate the representations of concept and non-concept examples in the embedding space. 

Although TCAV can be applied to all neural network classifiers, for simplicity we limit %explain this procedure for 
our experiments to binary RoBERTa-based abusive language classifiers. We choose the RoBerta-based models for their superior performance in processing social media data compared to other base language models \cite{liu2019roberta}. The concept, $C$, is defined by $N_C$ concept examples. Also, $N_R$ random examples are used to define non-concept examples. The RoBERTa representations for all these examples are calculated using $f_{emb}$, which maps an input text to its [CLS] token representation. Then, $P$ number of CAVs, $\upsilon_C^{p}$, are generated, each through training a linear classifier that separates a sub-sample (with size $N_c$) of concept examples from a sub-sample of random examples (with size $N_r$) in the RoBERTa embedding space. 
The \textit{conceptual sensitivity} of a label to the CAV, $\upsilon_C^{p}$, at input $x$ can be computed as the directional derivative $S_{C,p}(x)$:

\vspace{5pt}
$S_{C,p}(x) =
    \lim\limits_{\epsilon \to 0} \frac{h(f_{emb}(x)+\epsilon \upsilon_C^{p}) -h(f_{emb}(x))}{\epsilon}$

\begin{equation}
\quad \quad \quad = \bigtriangledown h(f_{emb}(x)).\upsilon_C^{p} 
\label{eq:sensitivity}
\end{equation}
 
\noindent where $h$ maps the RoBERTa representation to the logit value of the class of interest. 

In this work, we use two metrics to specify the influence of the concept on the model's prediction. First, we calculate $TCAV_{dir}$, the fraction of inputs in a set of input examples $X$, for which the directional derivative $S_{C,p}(x)$ is positive, i.e.:

\vspace{-4mm}
\begin{equation}
    TCAV_{dir}^{C,p} = \frac{| {x \in X:S_{C,p}(x)>0}|}{|X|}
\label{eq:TCAV_dir}
\end{equation}

 $TCAV_{dir}$ indicates the fraction of input examples for which the prediction scores of the model increase if the input representation is infinitesimally moved towards the concept representation. This metric has been widely used to identify if the label has learned the concept as an important signal for the label \cite{yeh2020completeness}.

Besides the widely used metric of $TCAV_{dir}$ (referred to as $TCAV$ score in previous work), we introduce a new metric, $TCAV_{mag}$, which considers the size of the directional derivatives, and measures the magnitude of the influence of the concept on the label for the positive directional derivatives: 
 
\vspace{-4mm}
\begin{equation}
    TCAV_{mag}^{C,p} = \frac{\sum_{x\in X,S_{C,p}(x)>0}^{ }S_{C,p}(x)}{|X|}
\label{eq:TCAV_mag}
\end{equation}

We demonstrate in our results that $TCAV_{mag}$ can be an indicator of the over-reliance of the label on the concept. 
When calculated for all CAVs, Equations \ref{eq:TCAV_dir} and \ref{eq:TCAV_mag} generate two distributions of scores with size $P$ for the concept $C$. Using a t-test, these distributions are compared with the distributions of $TCAV_{dir}$ and $TCAV_{mag}$ calculated for random examples to check for statistical significance \cite{kim2018interpretability}.

\begin{figure*}
    \centering
    \small
\includegraphics[width= 0.68\textwidth]{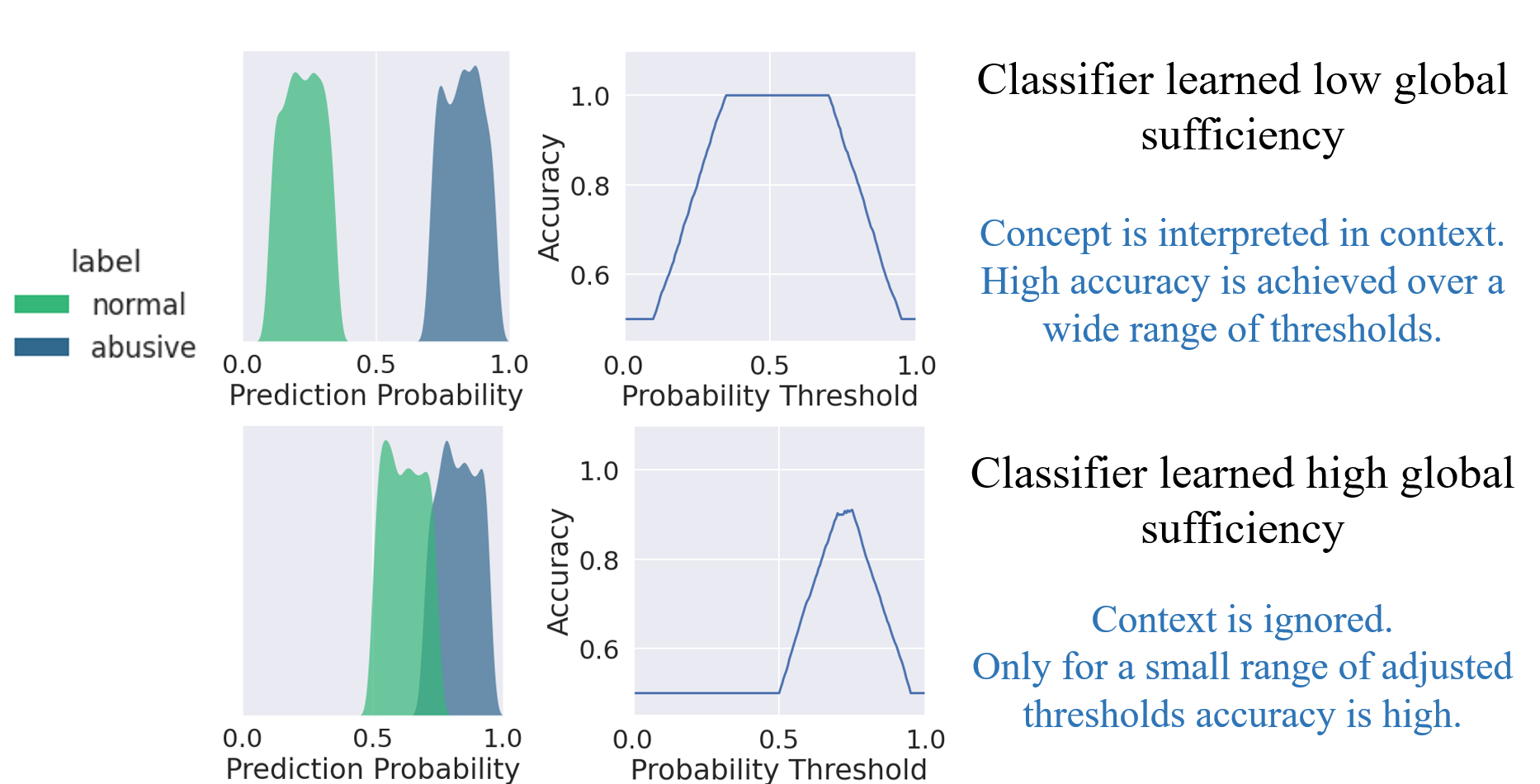}

\caption{Illustration of the potential distribution of probabilities generated by a trained binary classifier for a challenge set that represents an important concept, along with accuracy versus threshold curves. %The separability of the positive and the negative classes indicates the extent of the sufficiency the classifier has learned between the concept and the positive label.  
}
 \label{fig:illusteration}
\end{figure*}

% \begin{figure*}
% \centering
% \begin{tikzpicture}
%     % Image 1
%     \node[inner sep=0, anchor=north west] (image1) at (0,0) {\includegraphics[height=3cm]{images/1.png}};

%     % Image 2
%     \node[inner sep=0, anchor=north west, right=1cm of image1.north east] (image2) {\includegraphics[height=3cm]{images/2.png}};

%     % Image 3
%     \node[inner sep=0, anchor=north west, below=1cm of image1.south west] (image3) {\includegraphics[height=3cm]{images/3.png}};

%     % Image 4
%     \node[inner sep=0, anchor=north west, below=1cm of image2.south west] (image4) {\includegraphics[height=3cm]{images/4.png}};
% \end{tikzpicture}
% \caption{A 2x2 grid of aligned images}
% \label{fig:2x2_images}
% \end{figure*}

\section{False Global Sufficiency}
\label{sec:suffmetric}

Phenomenon P is considered globally sufficient for phenomenon Q ($P\Rightarrow Q$) if, whenever P occurs, Q also occurs \cite{zaeem2021cause}. In other words, global sufficiency refers to the extent to which a concept can explain the model's output across all instances in a held-out dataset, as opposed to the more studied topic of local sufficiency, which concerns the stability of an individual prediction for a given feature in perturbed contexts \cite{balkir-etal-2022-necessity}.

In a real-world setting, it is very unlikely that any single concept is truly sufficient for the label at a global level. In a binary classifier, a concept $C$ is falsely learned as sufficient for the positive label if all inputs containing $C$ are classified as positive by the classifier, regardless of context. This undesired dependency of the label on the concept suggests that the model has failed to learn how the concept interacts with context to influence the label. While this issue is closely related to spurious correlation, we use the term \textit{false global sufficiency} because spurious correlation typically implies that the feature is irrelevant to the label, and a correlation is learned due to a confounding factor. In contrast, we consider the cases where the feature is relevant and important but not globally sufficient.

To make this clearer, consider the case of abusive language detection and the concept of negative emotions; if the mere presence of negative emotions in a sentence always guarantees the prediction of the positive label (abuse), then the model has learned a false sufficiency relation between the concept and the label. It over-relies on this feature and ignores the context.

To quantify falsely learned global sufficiency, we consider two scenarios: 1) where a balanced challenge set is available, which contains $C$ in all of its examples (both classes), and 2) where no challenge set is available. For the first scenario we use the traditional approach of assessing accuracy of the classifier on a held-out test set. This approach provides a baseline in our evaluations. For the second scenario, we propose concept-based explanation metrics and compare them with the baselines obtained with the challenge sets.

\subsection{Quantifying the Falsely Learned Global Sufficiency with a Challenge Set} 
 \label{subsec:suff_challeng_set}

 Based on our definition of global sufficiency, one way to assess a model's over-reliance on a concept is to evaluate its performance on a held-out challenge set, $\mathbb{F}$, containing both positive and negative examples of the concept of interest \cite{yin2021towards}. For simplicity, we assume that this challenge set consists of equal numbers of positive and negative examples. If a model learns a high global sufficiency between the concept $C$ and the label of abuse, all examples in both positive and negative classes of a challenge set $\mathbb{F}$ will be labeled as abusive.  However, if the model interprets the concept in context, only the positive examples of $\mathbb{F}$ will receive the abusive label. This indicates that in cases where the decision threshold of the classifier clearly separates the probability distributions of the two classes, the model has learned a low global sufficiency between the concept and the label. 
%the model has learned a low global sufficiency for the concept, and its decision threshold clearly separates the probabilities of positive and negative classes. Therefore, we use the separation of probabilities by a decision threshold as an indicator of the low global sufficiency of the concept for the label. 

However, when comparing different classifiers, it is important to note that a reliable classifier should perform well (high precision and high recall) over a broad range of decision thresholds. This is because different applications may require different thresholds depending on the desired trade-off between precision and recall. For example, a classifier used to moderate social media content may need to prioritize precision over recall, which could mean using a high threshold to avoid false positives. On the other hand, a classifier used to detect all instances of abusive language may need to prioritize recall over precision, which would mean using a lower threshold to catch as many instances of abuse as possible, even if it means tolerating more false positives. Therefore, a classifier that is reliable over a wide range of decision thresholds can be more effective in different use cases, making it more practical and adaptable.

Figure~\ref{fig:illusteration} demonstrates two hypothetical cases for the distribution of probabilities that the classifiers might generate for the challenge set $\mathbb{F}$. A classifier that learned low global sufficiency between $C$ and the positive label generates easily separable distributions of probabilities for the positive and negative examples of $\mathbb{F}$. In other words, for a large range of decision thresholds, the two classes of $\mathbb{F}$ are separable, and high accuracy is achieved. Conversely, the classifier that has learned high global sufficiency between $C$ and the positive label assigns a similar distribution of probabilities to both negative and positive examples. The two classes of $\mathbb{F}$ are hardly separable, and for a wide range of thresholds, the accuracy is low. Note that in order for this classifier to be accurate, it requires a careful adjustment of the decision threshold with a labeled dataset. However, this process can be very costly.

Based on this discussion, we argue that $AUC\_Challenge$, the area under the curve of accuracy vs threshold, is a quantitative indicator of the separability of two classes of $\mathbb{F}$ for all decision thresholds. According to our definition above, global sufficiency is negatively correlated with the separability of these classes. Therefore, $False\_Suff$, described in Equation \ref{eq:suff_challeneg}, is a quantitative metric that can be used to compare the degree of sufficiency learned by the classifiers based on $\mathbb{F}$:
\compress
\begin{equation}
False\_Suff = 1-AUC\_Challenge%_{acc\; vs \; thr}
\label{eq:suff_challeneg}
\end{equation}

\begin{table*}
\centering
\small

\begin{tabular}{p{0.023\textwidth}p{0.30\textwidth}p{0.35\textwidth}p{0.1\textwidth}p{0.03\textwidth}}
\toprule

 & \textbf{Functionality} & \textbf{Example Test Case} & \textbf{Gold Label} & \textbf{n}  \\
\midrule
 & \textbf{F2}: Description using very negative attributes & ``{[}IDENTITY{]} are disgusting.'' \newline ``All {[}IDENTITY{]} like them are vile.'' & \hlred{ hateful } & 140 \\
& \textbf{F21}: Denouncement of hate that make direct reference to it & ``You have to stop calling {[}IDENTITY{]} disgusting.'' & \hlgreen{ non-hate } & 140  \\

%& \textbf{F10}: : Hate expressed using profanity & ``These [IDENTITY] should shut the f*ck up once and for all.'' & \hlred{ hateful } & 140 \\

% & \textbf{F18}: Neutral statements using protected group identifies & ``We are a group of {[}IDENTITY{]}'' \newline ``I live with two {[}IDENTITY{]}.'' & \hlgreen{ non-hate } & 126 \\

\bottomrule
\end{tabular}
\caption{\label{tab:hatecheck-cases}
Used test cases from HateCheck \citep{rottger2021hatecheck}. %We create a challenge set for the concept of \textit{``describing a protected group with negative attitudes''} by combining F2 and F21. F10 is combined with neutral sentences to create an easy set.  
}
\vspace{-1em} % Decrease the vertical space below the table
\end{table*}

\subsection{Quantifying the Falsely Learned Global Sufficiency with Concept-Based Explanations}
\label{subsec:suff_TCAV}

The practical application of the method detailed in Section \ref{subsec:suff_challeng_set} can be limited due to the necessity of creating a custom challenge set. In this section, we use concept-based explanation to measure the falsely learned global sufficiency in a scenario where a challenge set is not available, but a lexicon representing the concept of interest exists. Following the approach of \citet{nejadgholi2022towards}, we employ short templates and the concept lexicon to generate unlabeled concept examples. Then, we utilize the method described in Section \ref{sec:TCAV} to compute two metrics: $TCAV_{dir}$ and $TCAV_{mag}$. If the $TCAV_{dir}$ value for the concept significantly deviates from that of random concepts, it indicates that the classifier has learned an association between the label and the concept. A significant difference in $TCAV_{mag}$ compared to random concepts suggests a strong influence of the concept on the label, potentially causing the classifier to disregard the context when the concept is present. While the absolute values of these metrics might not be definitive, we show that they can be used to compare various classifiers in terms of the degree of global sufficiency they have learned for a concept.

\section{Sufficiency of the Concept of \textit{Describing Protected Groups with Negative Emotion} }
\label{sec:suff_neg_em}

In this section, we evaluate the metrics introduced in Section \ref{sec:suffmetric} in explaining the extent of the falsely learned sufficiency between a human-defined concept and the positive label of the classifiers. We specifically consider the concept of \textit{describing a protected group with negative emotion words} and refer to it as \textit{DesNegEm} for brevity. We chose this concept because it is tightly related to hate speech and is expected to be important for more general definitions of harmful language, such as toxic, abusive or offensive. Still, it is not a sufficient concept for these labels and has to be interpreted in the broader context (as shown by examples in Table~\ref{tab:hatecheck-cases}).

We consider three RoBERTa-based binary classifiers, publically available and trained with large English datasets. The models are trained for general definitions of abusive language, toxicity or offensive language. We refer to these classifiers by their training datasets: Jigsaw, Civil Comments (or Civil for brevity) and TweetEval. These models are described in detail in Appendix \ref{sec:model}

\begin{figure}
     \centering
     \includegraphics[trim={1cm 0 0 4cm}, clip,width=0.4\textwidth]{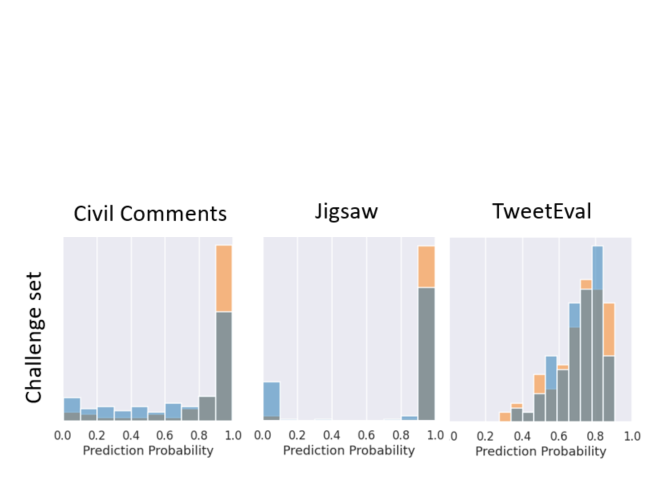}
    %\vspace{-10pt}
     \caption{Probability distributions generated by the classifiers for a challenge set (F2 and F21 of HateCheck).}
        \label{fig:all_suff}
         \vspace{-10pt}
\end{figure} 

\begin{figure}
     \centering
     \includegraphics[width=0.40\textwidth]{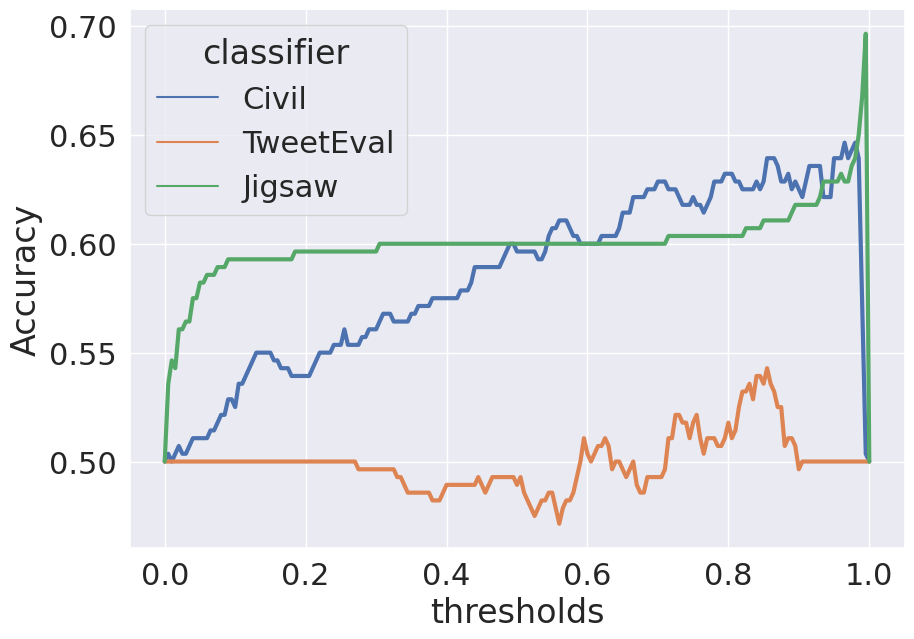}
     %\vspace{-10pt}
     \caption{Accuracy vs threshold curve for the challenge set (F2 and F21 of HateCheck).}
        \label{fig:all_auc}
        %\vspace{-10pt}
\end{figure}

\vspace*{0.3cm}
\noindent \textbf{Quantifying Sufficiency with a Challenge Set:} To calculate the metric described in Section \ref{subsec:suff_challeng_set}, we first use the HateCheck \citep{rottger2021hatecheck} test cases to build a challenge set for the concept of \textit{DesNegEm}. For that, we use the F2 and F21 functionalities of HateCheck, i.e., the hateful and non-hateful examples that include this concept (Table~\ref{tab:hatecheck-cases}). Figure~\ref{fig:all_suff} shows the distribution of probabilities that the three classifiers generate for this challenge set. We observe that, for a large range of decision thresholds, all three classifiers label the majority of the examples of both classes of the challenge set with a positive label. In other words, all three classifiers have learned a high sufficiency between \textit{DesNegEm} and the label of abuse. However, the extent of the learned sufficiency is different among the classifiers. The TweetEval classifier makes the least differentiation between the two classes and generates similar distributions of probabilities for negative and positive examples with the \textit{DesNegEm} concept. Because of this overlap between probability distributions of positive and negative classes, the accuracy of this classifier is low over all ranges of thresholds, as shown in Figure~\ref{fig:all_auc}. The false sufficiency learned by the Jigsaw and the Civil Comments classifiers is less extreme, and Jigsaw makes the most differentiation between the two classes.

\begin{table*}[]
\small
    \centering
    \begin{tabular}{c|c|c|c|c}
     & \multicolumn{2}{c|}{$TCAV_{dir}$}&\multicolumn{2}{c}{$TCAV_{mag}$} \\
     \hline
   classifier  & \multicolumn{1}{c|}{DesNegEm}& \multicolumn{1}{c|}{random}& \multicolumn{1}{c|}{DesNegEm}& \multicolumn{1}{c}{random}\\
     \hline
  \hline
 Civil &0.67(0.05)&0.5(0.4)&\textbf{0.05(0.03)}&0.00(0.01)\\
Jigsaw&\textbf{1(0)}&0.7(0.4)&0.04(0.01)&0.05(0.05)\\
TweetEval&\textbf{1(0)}&0.7(0.4)&\textbf{0.15(0.03)}&0.01(0.01)\\

  \hline
    \end{tabular}
    \caption{Mean and standard deviation of the TCAV score for explaining the sufficiency of \textit{Describing
Protected Groups with Negative Emotion} (DesNegEm) for the three classifiers. All scores statistically significantly different from random concepts are in boldface. }
    \label{tab:suff_TCAV}
    %\vspace{-1em} % Decrease the vertical space below the table
\end{table*}

This observation can be quantified with the $False\_Suff$ metric (Equation \ref{eq:suff_challeneg}) using the area under the curves in Figure~\ref{fig:all_auc}. We obtain $False\_Suff$ of $0.41$, $0.40$, and $0.50$ for Civil, Jigsaw and TweetEval, respectively. This metric shows a higher falsely learned sufficiency score for TweetEval than the two other classifiers, as expected. Based on these observations, we expect TCAV metrics to show lower scores for Civil and Jigsaw than the TweetEval classifier. 

\vspace*{0.3cm}
\noindent \textbf{Global Sufficiency with Concept-Based Explanations:} Here, we use the results obtained with the challenge set as a baseline to evaluate the TCAV-based metrics. Concept examples are generated using the template `\textit{<protected\_group> are <emotion\_word>.}', where \textit{<protected\_group>} is one of the protected groups \textit{women}, \textit{trans people}, \textit{gay people}, \textit{black people}, \textit{disabled people}, \textit{Muslims} and \textit{immigrants} as identified by \citet{rottger2021hatecheck}. For \textit{<emotion\_word>}, we use the \textit{disgust} and \textit{anger} categories of the NRC Emotion Intensity Lexicon (NRC-EIL) \cite{LREC18-AIL}. We use the NLTK package\footnote{\url{https://www.nltk.org/}} to filter out words other than adjectives, past tense verbs and past participles, and also remove the words with emotion intensity lower than $0.5$. After these steps, we are left with 368 concept words. We calculate the $TCAV_{dir}$ and $TCAV_{mag}$ scores for the concept of \textit{DesNegEm} and compare those to the metrics calculated for random concepts with t-test for statistical significance. For random concepts, the concept examples are random tweets collected with stop words. In our implementation of the TCAV procedure, $N_R = 1000$, $N_c= 50$, $N_r= 200$ and $N_C = 386$ (number of filtered lexicon words). For input examples, $X$, we use 2000 tweets collected with stop words.

As presented in Table~\ref{tab:suff_TCAV}, for the Civil classifier, $TCAV_{dir}$ is not significantly different from the random concept, indicating that the concept information might not always be encoded as a coherent concept in the embedding space of this classifier. However, $TCAV_{mag}$ is significantly higher than random, indicating that when the information is encoded well, the presence of this concept has a significant influence on the label of abuse. The other two classifiers have learned a strong association between the concept and the label, i.e., when the concept is added to a neutral context, the likelihood of the positive label increases.  However, only in the case of the TweetEval classifier, $TCAV_{mag}$ is significantly different from the random concepts, indicating a strong influence of the concept on the label, which might override the context. Therefore, for TweetEval the distribution of generated probabilities is mostly determined by the concept, not the context (similar distributions are obtained for the positive and negative examples of the challenge set). The other two classifiers consider the context to some extent and generate relatively different distributions of probabilities for the two classes.

\noindent \textbf{Discussion:} For all classifiers, the presence of the concept \textit{describing a protected group with negative emotion words} is a strong signal for the label of abuse. All classifiers struggle in considering the broader contexts in sentences such as `\textit{It is not acceptable to say <protected\_group> are disgusting.}' Among the three classifiers, TweetEval has learned a higher degree of sufficiency, leading to its worse performance on a challenge set containing this concept. The TCAV metrics can be used to compare the classifiers regarding the false sufficiency relationships they have learned. These metrics provide similar insights to what is learned from assessing global sufficiency with a challenge set.

\section{Global Sufficiency of Fine-Grained Negative Emotions Concepts}
\label{sec:suf_fine_grained}

In the previous section, we considered the concept of \textit{describing protected groups with negative emotions}, which is tightly related to hate speech, and thus prone to be mistakenly learned as sufficient for the label of abuse. In this section, we test our proposed method for a less obvious case by disentangling the concept of emotions and hate speech. We focus on the concept of \textit{describing a (non-protected) group of people with negative emotions}, which differs from the previous section in  1) removing the protected groups and replacing them with unprotected groups and 2) breaking down the emotion concept to more fine-grained levels. 

For fine-grained emotion concepts, we first develop a compact challenge set, examples of which are presented in Table~\ref{tab:emotion-cases}. Since we consider non-protected groups in this challenge set, the examples are labeled as abusive/non-abusive as opposed to hateful/non-hateful in HateCheck (shown in Table~\ref{tab:hatecheck-cases}). We assess the sufficiency of these concepts with the challenge set first and then compare the results to those of the proposed concept-based explanation metrics. Our goal is to investigate if the findings for the broad concept of describing protected groups with negative emotions can also be replicated at a more nuanced level of emotional granularity. We analyze the models for fine-grained categories of negative emotions, identified by \citet{LREC18-AIL}, namely \textit{disgust}, \textit{anger}, \textit{sadness}, and \textit{fear}. Similar pre-processing steps to what was described in Section \ref{sec:suff_neg_em} were performed to filter the lexicon in each category of emotions.

For the challenge set, we write five abusive and five non-abusive example templates for each emotion. Then we generate 40 abusive and 40 non-abusive examples by replacing <group> with one of the terms \textit{Canadians, Chinese people, doctors, teachers, school children, football players, my neighbours, and men} to represent non-protected groups.\footnote{Though nationality may be considered a protected characteristic in some contexts, we include ``Canadian'' and ``Chinese'' here since nationality was not included in HateCheck and therefore not covered in the previous section.} Full list of examples of this challenge set is available in our GitHub repository mentioned in Section \ref{sec:intro}. 
Equivalently, for the TCAV procedure for concept templates, we use `\textit{They are <emotion\_word>}', instead of `\textit{<protected\_group> are <emotion word>'}, which we used in Section \ref{sec:suff_neg_em}.

\begin{table*}
\centering
\small

\begin{tabular}{p{0.1\textwidth}p{0.55\textwidth}p{0.1\textwidth}p{0.03\textwidth}}
\toprule

  \textbf{Emotion} & \textbf{Example } & \textbf{Gold Label} \\
\midrule
 \textbf{disgust} & ``<group> are known for being \textbf{trash}.'' \newline ``It's reasonable that <group> are feeling \textbf{vindictive}.''&\hlred{abusive}\newline\hlgreen{non-abusive}\\
 \midrule
  \textbf{anger} & ``All <group> are \textbf{abhorrent}.'' \newline ``<group> were \textbf{infuriated} by the injustice. '' &\hlred{abusive}\newline\hlgreen{non-abusive}\\
\midrule
   \textbf{sadness} &  ``<group> are known for being \textbf{pathetic}.'' \newline ``<group> were \textbf{crushed} by the injustice.''&\hlred{abusive}\newline\hlgreen{non-abusive}\\

\midrule   
\textbf{fear} &   ``<group> are just so \textbf{scary}.'' \newline ``<group> are \textbf{threatened} by these policies.'' &\hlred{abusive}\newline\hlgreen{non-abusive}\\

\bottomrule
\end{tabular}
\caption{\label{tab:emotion-cases}
Challenge set created for emotion categories. In each sentence, we use a word from the corresponding category of  the NRC Emotion Intensity
Lexicon (NRC-EIL) \cite{LREC18-AIL}, shown in bolded font.}
%\vspace{-1em} % Decrease the vertical space below the table
\end{table*}

\begin{table*}[]
\small
    \centering
    \begin{tabular}{c|c|c|c|c||c|c|c|c}

      & \multicolumn{4}{c||}{$TCAV_{dir}$}& \multicolumn{4}{c}{$TCAV_{mag}$}\\%& trust& joy\\
      \cline{2-9}
        classifier  & disgust& anger& sadness& fear & disgust& anger& sadness& fear\\%& trust& joy\\
     \hline
  \hline
 Civil &0.48(0.37)&0.26(0.27)&0.31(0.31)&0.19(0.27)&0.05(0.06)&0.02(0.03)&0.02(0.03)&0.01(0.03)\\%&0(0)&0.02(0.06)\\
Jigsaw&\textbf{0.98(0.09)}&\textbf{0.93(0.2)}&0.91(0.2)&\textbf{0.95(0.18)}&0.08(0.03)&0.05(0.03)&0.03(0.02)&0.05(0.03)\\%&0.56(0.39)&0.48(0.4)\\
TweetEval&\textbf{1(0)}&\textbf{1(0)}&\textbf{1(0)}&\textbf{1(0)}&\textbf{0.20(0.04)}&\textbf{0.17(0.04)}&\textbf{0.11(0.03)}&\textbf{0.13(0.03)}\\%&0.9(0.2)&0.75(0.4)\\

\end{tabular}
\caption{Mean and standard deviation  of concept-based metrics for four negative emotion concepts. Scores that are significantly different from random concepts are in boldface. }
\label{tab:fine-grained_tcav_dir}
%\vspace{-1em} % Decrease the vertical space below the table
\end{table*}

\subsection{Results}

\begin{figure}
\centering
\includegraphics[width = 0.4\textwidth]{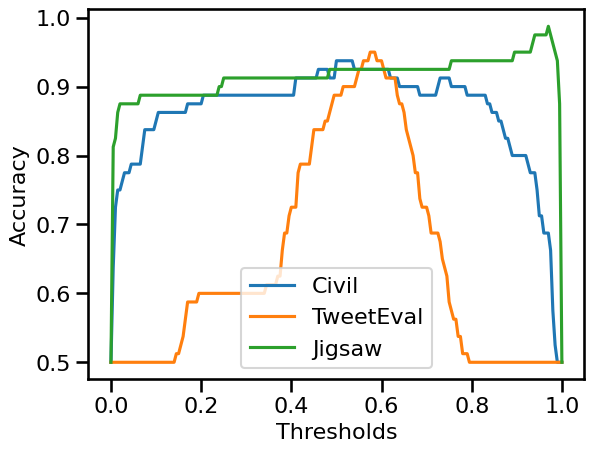}
\vspace{-10pt}
\caption{ Accuracy vs threshold for the disgust category of the challenge set.% TweetEval has the largest sufficiency or least separability for positive and negative examples of the challenge set. 
}
\label{fig:challenge_auc}
\vspace{-10pt}
\end{figure}

We first compare the three classifiers in handling negative emotions by investigating the results they produce for the challenge set. The $False_Suff$ scores in Table~\ref{tab:emotion_AUC} show that TweetEval has learned the highest sufficiency between these concepts and the label of abuse and therefore achieves the lowest separability between the positive and negative classes of the challenge set. To further clarify this we show the accuracy vs threshold curve for the \textit{disgust} category of the challenge set in Figure~\ref{fig:challenge_auc}. We observe that TweetEval only reaches high accuracies for a small range of thresholds, i.e, it generates a similar distribution of probabilities for the positive and negative classes that contain the emotion of \textit{disgust}. On the other hand, Jigsaw has learned the least global sufficiency and reaches high accuracy over a wide range of thresholds.  

Then we turn to the TCAV scores shown in Table~\ref{tab:fine-grained_tcav_dir}. First, $TCAV_{dir}$ shows that the Civil Comments classifier is not significantly sensitive to negative emotions, i.e., the feature of negative emotions is not fully learned as a coherent feature by this classifier. TweetEval, on the other hand, shows significant $TCAV_{dir}$ and $TCAV_{mag}$ scores, indicating that this classifier is not only sensitive to these concepts but the influence of the concept on the label is also significantly high. Jigsaw is the classifier that has learned the dependency between negative emotions and the label of abuse and therefore is sensitive to it (as indicated by $TCAV_{dir}$), but the magnitude of the influence of concept on the label is not significantly high, and the concept is interpreted in the larger context. Interestingly, the magnitude of the influence of \textit{disgust} and \textit{anger} is higher than \textit{fear} and \textit{sadness} for all classifiers, stating a higher association of \textit{disgust} and \textit{anger} with abusive language. These results are in line with conclusions drawn from assessing global sufficiency with a challenge set. 

\section{Related Works}
Most of the explainability works in NLP focus on feature importance methods to measure the importance of an input feature for the prediction at the local level \cite{bahdanau2015neural,sundararajan2017axiomatic, ribeiro2016should, lundberg2017unified}. However, recent works highlight that models should be assessed beyond feature importance criteria and that the reasoning behind the model's decisions should be investigated through explainability methods.
Some examples of such explainability methods include counterfactual reasoning \cite{wu2021polyjuice,kaushik2021explaining,ribeiro2020beyond,ross2020explaining} or necessity and sufficiency metrics \cite{balkir-etal-2022-necessity,joshi2022all}. Also, there is a need to compare various classifiers at the global level. Although local explanations can be aggregated to generate global explanations, they are usually obtained through costly interventions and are not practical to be applied on a large scale. For global explanations, a popular approach is to train probing classifiers \cite{conneau2018you}. However, probes only identify whether a classifier has learned a feature but stay silent about whether the feature is used in predictions \cite{belinkov2022probing,tenney2019you,rogers2020primer}. Amnesic probing is an extension of probing classifiers that identifies whether removing a feature influences the model's predictions, which relates to the notion of the global necessity of a human-understandable concept for a prediction \cite{ravfogel2020null, elazar2021amnesic}. 
%Also, probes don't give relative importance. 
Our work, on the other hand, focuses on the global sufficiency of concepts. While probing classifiers are applied to linguistic properties such as POS tagging, which are necessary for accurate language processing, we focus on human-defined semantic concepts that are known to be important for the label and test if they have been falsely learned as a sufficient cause for the label.

\begin{table}[]
    \centering
    \small
    \begin{tabular}{c|cccc}
     & \multicolumn{4}{c}{$False\_Suff$} \\
     \hline
  classifier    & disgust & anger &sadness&fear\\
  \hline
  Civil &0.13&0.35&0.19&0.25\\
  Jigsaw&0.08&0.28&0.14&0.22\\
  TweetEval&0.36&0.36&0.35&0.35\\
  \hline
    \end{tabular}
    \caption{The global sufficiency of emotion categories learned by classifiers with respect to the challenge set described in Table~\ref{tab:emotion-cases}. }
    \label{tab:emotion_AUC}
    %\vspace{-1em} % Decrease the vertical space below the table
\end{table}

Concept-based explanations have been introduced in computer vision and are mostly used to explain image classification models \cite{graziani2018regression,  ghorbani2019towards,yeh2020completeness}. %\citet{ghorbani2019towards} states that a concept needs to satisfy the properties of \textit{meaningfulness}, \textit{coherency} and \textit{importance} for the task at hand. 
% Some examples of concepts in computer vision tasks are the concept of \textit{stripes} for the class of \textit{zebra} \cite{kim2018interpretability}, the concept of \textit{white coat} for the class of \textit{doctor} \cite{pandey_2021}, and the concept of \textit{nuclei texture} in the detection of tumour tissue in breast lymph node samples \cite{graziani2018regression}. 
% [ADD literature on causal concept-based explanations. ] 
In NLP, concept-based explanations were used to measure the sensitivity of an abusive language classifier to the emerging concept of \textit{COVID-related anti-Asian hate speech} \cite{nejadgholi-etal-2022-improving}, to assess the fairness of abusive language classifiers in using the concept of sentiment  \cite{nejadgholi2022towards}, and to explain a text classifier with reference to the concepts identified through topic modelling \citep{yeh2020completeness}. To the best of our knowledge, our work is the first that uses concept-based explanations to assess the sufficiency of human-defined concepts in text classification.

\section{Conclusion}
Concept-based explanations can assess the influence  of a concept on a model's predictions. 
%We used two metrics; one identifies whether the classifier has learned an association between a concept and a label, and the other measures the magnitude of the influence of the concept on the label. For the task of abusive language classification, we showed classifiers that learned a high association between negative emotions and abuse but did not over-rely on this concept because of its large magnitude of influence, could reach the best performances.  
We used two metrics based on the TCAV method: the TCAV \textit{direction} score identifies whether the classifier has learned an association between a concept and a label, and the TCAV \textit{magnitude} score measures the extent of the influence of the concept on the label. We showed that the best-performing abusive language classifiers learned that negative emotion is associated with abuse (positive direction) but did not over-rely on this concept (low magnitude); that is, they did not overestimate the global sufficiency of that concept.

Our method can potentially be used for other NLP classification tasks. This approach is suitable for tasks where certain concepts are closely related to the label, but not enough to make a definitive determination. For example, in sentiment analysis, the price of products may have a strong connection to negative sentiment, but is insufficient to determine it. Further research should explore how concept-based explanations can help identify cases where certain concepts are relied upon too heavily in abusive language detection or other NLP classification tasks.

\section{Limitations}
Our work has limitations. First, we use the TCAV framework, which assumes that concepts are encoded in the linear space of semantic representations. However, recent works show that in some cases, linear discriminants are not enough to define the semantic representations of concepts in the embedding spaces \cite{koh2020concept}. Future work should consider nonlinear discriminants to accurately represent concepts in the hidden layers of NLP neural networks. 

In this study, we used simple challenge sets to obtain a baseline for assessing the effectiveness of concept-based explanations in measuring false global sufficiency. Future work should focus on curating challenge sets by annotating user-generated data for the label and the concepts, in order to achieve a stronger baseline.

Our work is limited to pre-defined concepts and requires human input to define the concepts with examples. However, defining concepts in TCAV is less restrictive than pre-defining features in other explainability methods, in that concepts are abstract ideas that can be defined without requiring in-depth knowledge of the model's inner workings or the specific features it is using. This allows for a more flexible approach where users can test the model regarding their concept of interest.

 Our method can only be applied to concepts that are known to be important for the classifier and are prone to being over-represented in training sets. %we expect a low false negative rate for one of our metrics. 
It's important to check this condition independently before using our metrics. In cases where this condition does not hold true, the metrics we use in our work may be interpreted differently and may not be reliable indicators of global sufficiency. Also, we only considered two variations of emotion-related concepts. Other variations such as \textit{expression of negative emotions by the writer of the post} should be investigated in future work.

Further, our metrics are limited to cases where different classifiers are being compared since the most important information is in the relative value of the metrics. Our metrics should not be used as absolute scores for testing a classifier. 

Testing a classifier for false causal relationships is most valuable for detecting the potential flaws of the models. If our metrics do not reveal a false relationship between the concept and the label, that should not be interpreted as an indicator of a flawless model.

\section*{Ethical Statement}

% In the current study, we assessed models for learning hateful concepts toward protected groups. We included several protected groups, but the list is not exhaustive. More protected groups should be included in the future. 
% Additionally, it is known that the label used to refer to a social group can itself communicate bias (consider, for example, the difference between \textit{immigrants} versus \textit{migrants} versus \textit{expats}) \citep{beukeboom2019stereotypes}. We have not analyzed the effect of this form of bias on the explanations here.
% Furthermore, other legally non-protected groups (e.g., based on physical appearances, education, etc.) should also be considered as we strive toward inclusive and safe online spaces. 

As with most AI technology, this approach can be used adversely to exploit the system's vulnerabilities and produce toxic texts that would be undetectable by the studied classifier. 
Specifically, for methods that require access to the model's inner layers, care should be taken so that only trusted parties could gain such access. The obtained knowledge should only be used for model transparency purposes, and the security concerns should be adequately addressed. 

Regarding environmental concerns, contemporary NLP systems based on pre-trained large language models, such as RoBERTa, require significant computational resources to train and fine-tune. Larger training datasets, used for fine-tuning, usually result in better classification performance but also an even higher computational cost. To lower the cost of this study and its negative impact on the environment, we chose to use existing, publicly available classification models.

\bibliography{anthology,custom}
\bibliographystyle{acl_natbib}

\appendix
% \section{Sufficiency and training datasets}
\section{Models}
\label{sec:model}
We include the following publicly available abusive language  classification models in this study:

\begin{itemize}

    \item %\href{https://huggingface.co/SkolkovoInstitute/roberta_toxicity_classifier/tree/main}{roberta\_toxicity\_classifier}, a binary classifier to identify toxic and non-toxic language trained on Jigsaw
    \textbf{Jigsaw}\footnote{\url{https://huggingface.co/SkolkovoInstitute/roberta_toxicity_classifier/tree/main}}: a RoBERTa-based binary toxicity classifier fine-tuned on the combination of two datasets created by Jigsaw and used in Kaggle competitions on toxicity prediction in 2018-2020. The first dataset, Wikipedia Toxic Comments \citep{wulczyn2017}, includes 160K comments from Wikipedia talk pages. The second dataset, Civil Comments \citep{borkan2019}, comprises over 1.8M online comments from news websites. Both datasets are annotated for toxicity (and its subtypes) by crowd-sourcing. The model creators report the AUC of 0.98 and F1-score of 0.76 on the Wikipedia Toxic Comments test set. The model is released under CC BY-NC-SA 4.0.  

    \item 
    %\href{https://huggingface.co/unitary/unbiased-toxic-roberta}{A multi-class classifier} to identify various types of abusive language, including identity-hate, insult and sexually explicit trained on Civil comments.

    \textbf{Civil Comments}\footnote{\url{https://huggingface.co/unitary/unbiased-toxic-roberta}} \citep{Detoxify}: a multi-class RoBERTa-based model fine-tuned on the Civil Comments dataset to predict toxicity and six toxicity subtypes (severe toxicity, obscene, threat, insult, identity attack, and sexual explicit). A part of the dataset is annotated for identity groups targeted in toxic comments. The prediction model is trained to optimize the outcome fairness for the groups in addition to the overall accuracy. This is achieved through the loss function that combines the weighted loss functions for two tasks, toxicity prediction and identity prediction \citep{hanu_2020}. 

        \item 
    %\href{https://huggingface.co/cardiffnlp/twitter-roberta-base-offensive}{RobertaBaseTweetEval}, a binary-class classifier to identify offensive, non-offensive language  trained on %\href{https://huggingface.co/datasets/tweet_eval}
    %{tweetEval}
    \textbf{TweetEval}\footnote{\url{https://huggingface.co/cardiffnlp/twitter-roberta-base-offensive}} \citep{barbieri-etal-2020-tweeteval}: a RoBERTa-based binary classifier to detect offensive language, released as part of the TweetEval evaluation benchmark. The model was trained on 58M tweets and then fine-tuned on the Offensive Language Identification Dataset (OLID) \citep{zampieri-etal-2019-predicting}. The OLID training set comprises about 12K tweets. The model achieved the macro-averaged F1-score of 77.1 on the OLID test set.

    % \item Wikipedia toxicity, a binary classifier trained on WikiToxicity  
    
    % \item A three-class classifier to identify abusive, hateful and neutral language trained on Founta 

\end{itemize}

\end{document}